\documentclass[journal]{IEEEtran}
\usepackage[utf8]{inputenc}
\usepackage{graphicx}
\usepackage{amsmath} 
\usepackage{graphicx}

\usepackage{float}
\usepackage{booktabs}
\usepackage{multirow}

\usepackage[nocompress]{cite}
\ifCLASSINFOpdf
\else
\fi

\begin{document}
%
\title{Limited-View Photoacoustic Imaging Reconstruction Via High-quality Self-supervised Neural Representation}
%
%
%

\author{Youshen~xiao,
        Yuting~Shen,
        Bowei~Yao,
        Xiran~Cai,
        Yuyao~Zhang,
        and~Fei~Gao

\thanks{ShanghaiTech University, Shanghai 201210, China.}
\thanks{gaofei@shanghaitech.edu.cn}
}

\markboth{Journal of \LaTeX\ Class Files,~Vol.~14, No.~4, July~2024}%
{Shell \MakeLowercase{\textit{et al.}}: Bare Demo of IEEEtran.cls for IEEE Journals}
%



\maketitle

\begin{abstract}
In practical applications within the human body, it is often challenging to fully encompass the target tissue or organ, necessitating the use of limited-view arrays, which can lead to the loss of crucial information. Addressing the reconstruction of photoacoustic sensor signals in limited-view detection spaces has become a focal point of current research. In this study, we introduce a self-supervised network termed HIgh-quality Self-supervised neural representation (HIS), which tackles the inverse problem of photoacoustic imaging to reconstruct high-quality photoacoustic images from sensor data acquired under limited viewpoints. We regard the desired reconstructed photoacoustic image as an implicit continuous function in 2D image space, viewing the pixels of the image as sparse discrete samples. The HIS's objective is to learn the continuous function from limited observations by utilizing a fully connected neural network combined with Fourier feature position encoding. By simply minimizing the error between the network's predicted sensor data and the actual sensor data, HIS is trained to represent the observed continuous model. The results indicate that the proposed HIS model offers superior image reconstruction quality compared to three commonly used methods for photoacoustic image reconstruction.
\end{abstract}

\begin{IEEEkeywords}
photoacoustic, limited-view, implicit neural representation, self-supervised
\end{IEEEkeywords}

%
\IEEEpeerreviewmaketitle

\section{Introduction}
%
%
%
%
\IEEEPARstart{P}{hotoacoustic} tomography (PAT), as an emerging non-invasive medical imaging technique, has garnered widespread attention in recent years, showcasing promising preclinical and clinical applications\cite{1,2,3,4}. This technology combines the high contrast of optical imaging and the deep penetration of ultrasound imaging through the photoacoustic effect, making it a highly promising imaging modality. Laser pulses stimulate the emission of ultrasound signals from biological tissues, which are then detected by ultrasound transducers. The intensity and profile of the photoacoustic signals are directly related to tissue optical absorption, which in turn is closely linked to the physiological and pathological states of the tissue. By reconstructing the photoacoustic signals, the distribution of radiation absorption within tissues can be depicted, providing valuable information for the medical field\cite{5,6,7}.
\par However, in practical applications, it is often challenging to fully encompass the target tissue or organ, necessitating the use of limited-view arrays, which may result in the loss of important information. Addressing the reconstruction problem of PAT signals sampled in a limited-view detection space has become a focal point of current research. Traditional PAT reconstruction algorithms can be broadly categorized into two types: linear reconstruction methods and model-based reconstruction methods\cite{8,9,10}. Linear reconstruction methods such as back-projection and time reversal, while computationally efficient, often exhibit distorted images with artifacts, especially under limited-view views. In contrast, model-based reconstruction methods rely on optimized iterative strategies to minimize the difference between the predicted signals estimated by the photoacoustic forward model and the measured signals. To achieve accurate reconstruction, model-based algorithms\cite{11} require precise model matrices and appropriate prior knowledge. While incorporating prior knowledge can reduce artifacts and significantly improve reconstruction quality, there is still a lack of effective regularization term expressions for a comprehensive description of the reconstruction results, potentially leading to suboptimal outcomes. Additionally, model-based reconstruction algorithms are time-consuming, and the weighting of regularization terms significantly influences the reconstruction results, limiting the method's performance. Therefore, overcoming the impact of limited-view detection on PAT reconstruction to enhance the accuracy and stability of reconstruction results remains a critical challenge that needs to be addressed in current research.
\par In recent years, significant progress has been made in the field of biomedical image processing with the advent of deep learning-based methods\cite{12,13,14,15}. Many deep learning approaches have been widely applied in image reconstruction, image post-processing, image segmentation, and lesion classification, including PAT reconstruction. Within the realm of photoacoustic tomography, deep learning-based methods primarily encompass post-processing techniques that utilize networks to eliminate artifacts present in images obtained through conventional analytical methods. Researchers are actively exploring various deep learning architectures to enhance image resolution, reduce noise and image artifacts, achieve higher-quality reconstruction imaging, and reconstruct images from limited-view data\cite{16,17}. Currently, detection angles of 70°\cite{Guo2024ScorebasedGM}, 90°, 120°\cite{18}, and 180° are commonly used in experimental limited-view settings(Fig. 1).

\begin{figure}[htp]
    \centering
    \includegraphics[width=4cm]{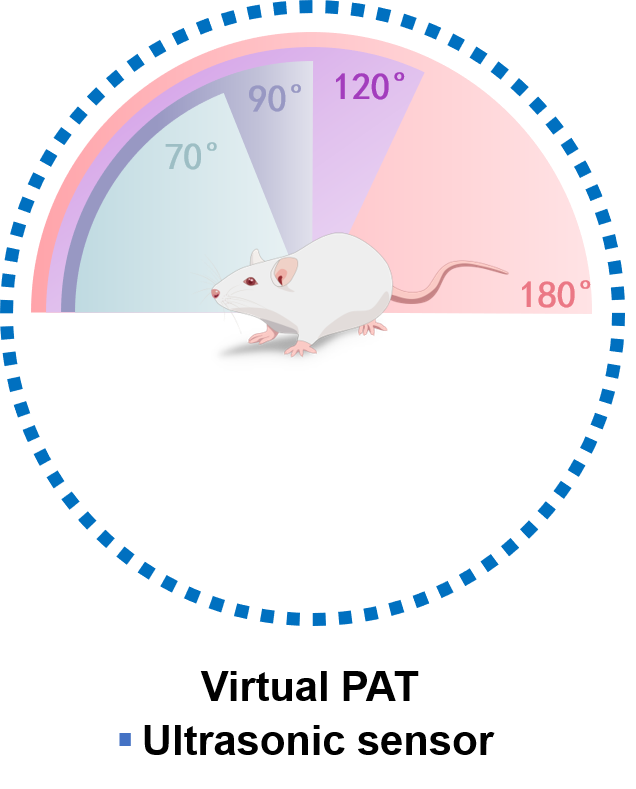}
    \caption{Commonly encountered geometric illustrations in experiments include 360° full-range tomographic imaging, along with representative angles such as 180°, 120°, 90°, and 70°.}
    \label{fig:galaxy}
\end{figure}

Numerous studies have indicated that applying deep learning methods can enhance photoacoustic computed tomography reconstruction using circular arrays at various limited viewing angles. For instance, Guan et al\cite{17}. introduced a deep learning approach called "Pixel-DL" for reconstructing vessel-simulated data from a 180° detection angle, which improves the image reconstruction quality for limited-view and sparse photoacoustic imaging techniques. However, a challenge exists in that the reconstructed images may contain additional vessels not present in the ground truth images. On the other hand, Tong et al\cite{20}. proposed a feature projection network named “FPnet”, which achieves higher reconstruction quality from a 270° limited viewing angle.

\par These supervised deep learning reconstruction methods mainly leverage convolutional neural networks (CNNs) to learn an end-to-end mapping from low-quality images or their signals acquired at limited viewing angles in large datasets to corresponding high-quality full-view images. However, deep learning methods heavily rely on the data distribution of image pairs in the training dataset, where larger training datasets typically offer improved performance. Nevertheless, generalization issues are significant due to disparities between different training datasets, such as variations in limited-view under-sampling schemes, sensor quantities, and different organ sites, which significantly impact the performance of the trained networks.

\par A recent method known as Implicit Neural Representations (INR)\cite{21,22} has been introduced, which utilizes coordinate-based deep neural networks in a self-supervised manner to model and represent three-dimensional scenes from sparse two-dimensional views. Unlike traditional approaches, the signal values in INR are discretely stored on a coordinate grid. The key innovation of this new method lies in training neural networks equipped with continuous non-linear activation functions to approximate the complex relationship between coordinates and corresponding signal values, ultimately providing a continuous signal representation. At the core of INR are continuous implicit functions parameterized by Multi-Layer Perceptrons (MLPs). INR has garnered significant attention for its ability to more compactly and flexibly learn tasks involving complex and high-dimensional data. They demonstrate immense potential in applications such as computer graphics, view synthesis, and image super-resolution.

\par In this paper, we have introduced the HIgh-quality Self-supervised neural representation (HIS) method, which is capable of reconstructing high-quality, artifact-free photoacoustic images from various common limited viewing angles without the need for external data. Setting itself apart from traditional CNN architectures, we have incorporated an INR network for learning and representing the initial acoustic pressure of the desired reconstructed images. Specifically, we hypothesize the desired reconstructed image as an implicit continuous function of two-dimensional image space coordinates, treating the signals collected by sensors at limited-view as sparse discrete samples of this function after forward propagation. Subsequently, utilizing an extended MLP, we consider the 2D coordinates $(x, y)$ of the position encoding as inputs to the network, which then outputs the image intensity $I(x, y)$ at that position. By minimizing the error between the network predictions and the acoustic pressure signals collected by each sensor at limited-view, HIS is trained to reconstruct a continuous model of the observed photoacoustic image.

\par We have evaluated the proposed method on samples ranging from simple to complex structures, and both qualitative and quantitative results indicate that HIS outperforms current common reconstruction methods in photoacoustic tomography. To the best of our knowledge, HIS is the first method to apply INR to the reconstruction of photoacoustic images from limited viewing angles. The primary advantages of HIS can be summarized as follows:

\begin{enumerate}
  \item Our proposed HIS method is capable of recovering high-quality photoacoustic images from acoustic pressure signals acquired at limited-view without the need for external data.
  \item Compared to traditional model-based methods, HIS significantly enhances the speed of image reconstruction.
\end{enumerate}

\section{Principles and methods}

\subsection{The model-based iterative method of PAT}
In photoacoustic imaging (PAT), tissues are exposed to short-pulsed laser irradiation. The tissues absorb part of the optical energy and convert it into thermal energy. This rapid temperature increase results in thermal expansion, generating ultrasound, also known as the photoacoustic signal, that propagates outward. The photoacoustic signal is captured by an ultrasound transducer and then undergoes data processing to reconstruct the initial pressure distribution. In PAT, the calculation formula for the initial acoustic pressure\cite{23} is typically represented by Equation (1):
\begin{equation}p_0=\Gamma\eta_{th}\mu_aF\end{equation}

\begin{figure*}[htbp]
  \centering
  \includegraphics[width=\linewidth]{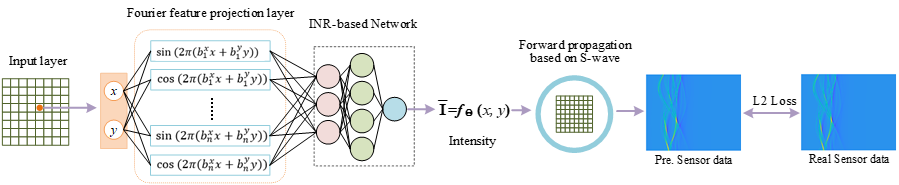}
  \caption{Workﬂow of the proposed HIS model. The network parameterizing implicit function $f_{\Theta}$ takes the coordinate p of sampling points as input and predicts the image intensity $I=f_{\Theta}({p})$at these positions.Then,the sensor data $\bar{\hat{\mathbf{y}}}_{s}$ obtained from the predicted image are calculated by the forword operator. Finally, we optimize the network by minimizing the loss between the predicted sensor data $\bar{\hat{\mathbf{y}}}_{s}$ and real sensor data $\mathbf{y}_s$ from acquired limited-view.} 
  \label{fig: Figure 2}
\end{figure*}
\begin{figure*}[htbp]
  \centering
  \includegraphics[width=13cm]{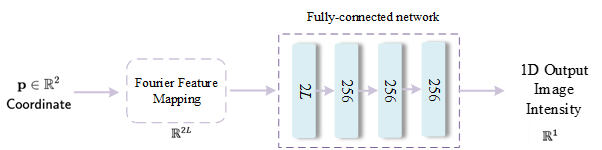}
  \caption{The architecture of the neural network used for parameterizing the implicit function $f_{\Theta}$, which consists of the fourier encoding and a three-layers MLP. }
  \label{fig: Figure 2}
\end{figure*}

where, $\Gamma$ represents the Gruneisen parameter (dimensionless), $\eta_{th}$ denotes the percentage of energy converted into thermal energy, $\mu_a$ is the optical absorption coefficient, and $F$ stands for the optical fluence. The generation and propagation of the photoacoustic wave are typically described by the general photoacoustic equation, which can be expressed as Equation (2):
\begin{equation}\left(\nabla^2-\frac{1}{\nu_s^2}\frac{\partial^2}{\partial t^2}\right)p(r,t)=-\frac{\beta}{k\nu_s^2}\frac{\partial^2T(r,t)}{\partial t^2}\end{equation}
where, $p(r,t)$ represents the instantaneous acoustic pressure at a given position $r$ and time $t$, $\nu_s$ is the speed of sound in water, $k$ denotes the isothermal compressibility, and $T$ represents the temperature rise. By employing the Green's function method, the solution for $p(r,t)$ can be obtained, as shown in Equation (3).
\begin{equation}p(r,t)=\frac{1}{4\pi\nu_{s}^{2}}\frac{\partial}{\partial t}\left[\frac{1}{\nu_{s}t}\int dr'p_{0}(r')\delta\left(t-\frac{|r-r'|}{\nu_{s}}\right)\right]\end{equation}
where, $r'$ denotes the position of the acoustic source, $p_{0}(r')$ represents the acoustic pressure at the source location, and also signifies the initial pressure distribution.
\par A matrix $x$ was initially characterized as the initial pressure distribution. Subsequently, this matrix $x$ undergoes transformation into the time-domain photoacoustic signal captured by sensors. The captured signal, denoted as $y$, is subject to the effects of sampling conditions and environmental factors, conforming to Equation (4):
\begin{equation}y=Ax\end{equation}
where $A$ symbolizes the forward operator in PAT, typically realized through the utilization of the k-Wave toolbox\cite{24}. 
\par In this work, we opt to employ the k-Wave approach for computing the forward operator.

\subsection{Image Reconstruction Strategies}
In the proposed HIS model, we represent the desired photoacoustic image $x$ as a continuous function parameterized by a neural network:
\begin{equation}I=f_\Theta( {p})\end{equation}

Where $\Theta$ represents the trainable parameters of the network (weights and biases), $\mathbf{p}=(x,y)\in {R}^{2}$ denotes any two-dimensional spatial coordinates within the imaging plane, $I\in {R}$ represents the image intensity (initial pressure) corresponding to position $p$ in the image $x$. Leveraging the acquired sensor data from a limited-view, we employ the backpropagation gradient descent algorithm to optimize the network towards approximating the implicit function, aiming to minimize the following objective:
\begin{equation}\widehat{\boldsymbol{\Theta}}=\arg\min_{\boldsymbol{\Theta}}\mathcal{L}(\hat{\mathbf{y}}_s,\mathbf{y}_s),\mathrm{with}\hat{\mathbf{y}}_s=\mathbf{A}f_{\boldsymbol{\Theta}}\end{equation}

where $\hat{\mathbf{y}}_s$ represents the predicted sensor data and $L$ is the loss function that measures the discrepancy between the predicted sensor data $\hat{\mathbf{y}}_s$ and the acquired sensor data ${\mathbf{y}}_s$.
\par Once the optimization is converged, the high-quality photoacoustic image $x$ can be reconstructed by feeding all the spatial coordinates $p$ into the MLP $f_{\Theta}$  to predict the corresponding intensity values $I$. The workﬂow of the proposed HIS model is shown in Fig. 2.

\subsection{Learning The Implicit Function}
Fig. 2 illustrates the process of learning an implicit function through a neural network. Given a sensor data $\mathbf{y}_s\in {R}^{K\times M}$ under limited viewing angles, where $K$ and $M$ are the number of sensors and the sampling points for each sensor respectively, we ﬁrst reshape $\mathbf{y}_{s}$ into a column vector of $(K\times M)\times1$, then similarly reshape the input image $x$ into a column vector.
\par As the summation operator (4) is differentiable, the neural network employed to parameterize the implicit function $f_{\Theta}$ can be optimized using the back-propagation gradient descent algorithm to minimize the discrepancy between the predicted sensor data $\mathbf{\hat{y}}_{s}$ and the actual sensor data $\mathbf{y}_{s}$ obtained from limited-view. In this study, We use an L2-loss between the real sensor data and predict sensor data to train in the image prediction module.

\subsection{Network Architecture}
As illustrated in Fig. 3, HIS consists of a position encoding section (via Fourier feature mapping), and a fully-connected network (MLP). It takes 2D spatial coordinate as input and outputs the intensity of the image at that location.

\begin{enumerate}
  \item $Fourier Encoding$: In theory, a multilayer perceptron can be utilized to approximate any complex function\cite{26}. However, recent studies have indicated that deep learning networks tend to learn lower-frequency functions during practical training, as highlighted by Rahaman et al. in their research\cite{27}. To address this issue, Mildenhofer et al. introduced the concept of Fourier feature mapping[28], which involves mapping low-dimensional inputs to higher-dimensional spaces, enabling the network to capture higher-frequency image features. This approach provides an effective way for deep learning networks to learn high-frequency image features, opening up new possibilities in the advancement of image processing and computer vision. In HIS, before feeding the 2D coordinates into the fully-connected network, we apply Fourier feature mapping\cite{28} to transform them to a higher-dimensional space ${R}^{2L}$ (2L $>$ 2). Here, $\gamma(\cdot)$ represents the Fourier feature mapping function from the space ${R}^2$ to ${R}^{2L}$, and it is computed as follows:
  $$\gamma(\mathrm{p})=[\cos{(2\pi B{p})},\sin{(2\pi B\mathrm{p})}]^T$$
  where $\mathrm{p}=(x,y)\in {R}^{2}$ and each element in $B\in {R}^{L\times2}$ is sampled from gaussian distribution $N\left(0,1\right)$.
  \item $Three-Layers MLP$: Following the fourier encoding process, the 2D input coordinate  is transformed into the high-dimensional feature vector . Subsequently, a three-layer Multilayer Perceptron (MLP) is employed to convert the feature vector  into the image intensity $I$. The two hidden layers within the MLP consist of 256 neurons each, accompanied by ReLU activation functions, while the output layer is appended with a Sigmoid activation function.
\end{enumerate}

\subsection{Dataset}
This work utilizes both simulated and experimental datasets. In order to obtain a sufficient number of simulated  datasets under full-view sampling, a virtual PAT was constructed based on the k-Wave toolbox\cite{24}.The k-Wave toolbox is widely utilized in photoacoustic tomography. Virtual photoacoustic tomography (PAT) enables the forward process of PAT imaging under arbitrary projections. The entire computational domain is set to 50 × 50 mm, with a total grid size of 440 × 440. The number of ultrasound transducers in the region of interest is determined by experimental requirements, typically set to 512 to ensure basic ground truth acquisition.The ultrasonic transducers are placed at a radius of 22 mm from the center of the grid. The surrounding medium is water with a density of 1000 $\mathrm{kg/m^3}$. The speed of sound is set to 1500 $\mathrm{m/s}$.
\par In this study, we first created three sets of simulated data with varying complexity, aiming to preliminarily validate the reconstruction performance of our HIS model under limited-view. To further validate the reconstruction capability of our model under practical limited-view, we also performed reconstructions on a circular phantom and in vivo experimental data of mice's abdomens. The circular phantom and in vivo experimental data of mice's abdomens were obtained from \cite{Davoudi2019DeepLO}.

\subsection{Training Parameters}
We adopt Adam optimizer to minimize the L2-loss function and the hyper-parameters of the Adam are as follows: $\beta_{1}=0.9$, $\beta_{2}=0.999$, $\epsilon=10^{-8}$. Commencing with an initial learning rate of $10^{-4}$. The entirety of training spans 10000 epochs, a task efficiently accomplished in approximately 4 minutes utilizing a single GeForce GTX 1080Ti GPU.

\section{Results}
\subsection{Results of simulation data}
To validate the efficacy of the HIS model, we conducted experiments using simulated data. Specifically, we designed three sets of simulation data, varying in complexity, and employed the K-Wave to simulate PAT acquisitions with limited-view views at 180°, 120°, 90°, and 70°. This comprehensive approach aimed to systematically assess the model's performance across a range of scenarios.
\par The proposed method was benchmarked against established techniques, including UBP, TR, and MB, with performance comparisons depicted in Figures 4 through 6. 
\par Figure 4 illustrates the outcomes for simulations involving simple geometric phantom, revealing that both Universal back-projection(UBP), Time Reversal(TR) and model-based(MB) methods are prone to severe artifact generation under limited-view conditions, particularly at more extreme angular constraints. The iterative MB approach manages to effectively suppress artifacts at a 180° view, but its performance deteriorates as the angular coverage decreases.
\par In contrast, the HIS model demonstrates superior capability in mitigating artifact formation, successfully reconstructing the geometry of the phantom with high fidelity at 180°, 120°, and 90° viewpoints. Even when confronted with the stringent challenge of a 70° limited view – where some structural details inevitably remain unresolved – the HIS method significantly outperforms its counterparts by accurately reconstructing the majority of the phantom’s structure. This underscores the robustness and effectiveness of the HIS model across a broad spectrum of viewing angles, especially in scenarios where data scarcity poses a significant hurdle to accurate image reconstruction.
\begin{figure}[htp]
    \centering
    \includegraphics[width=9cm]{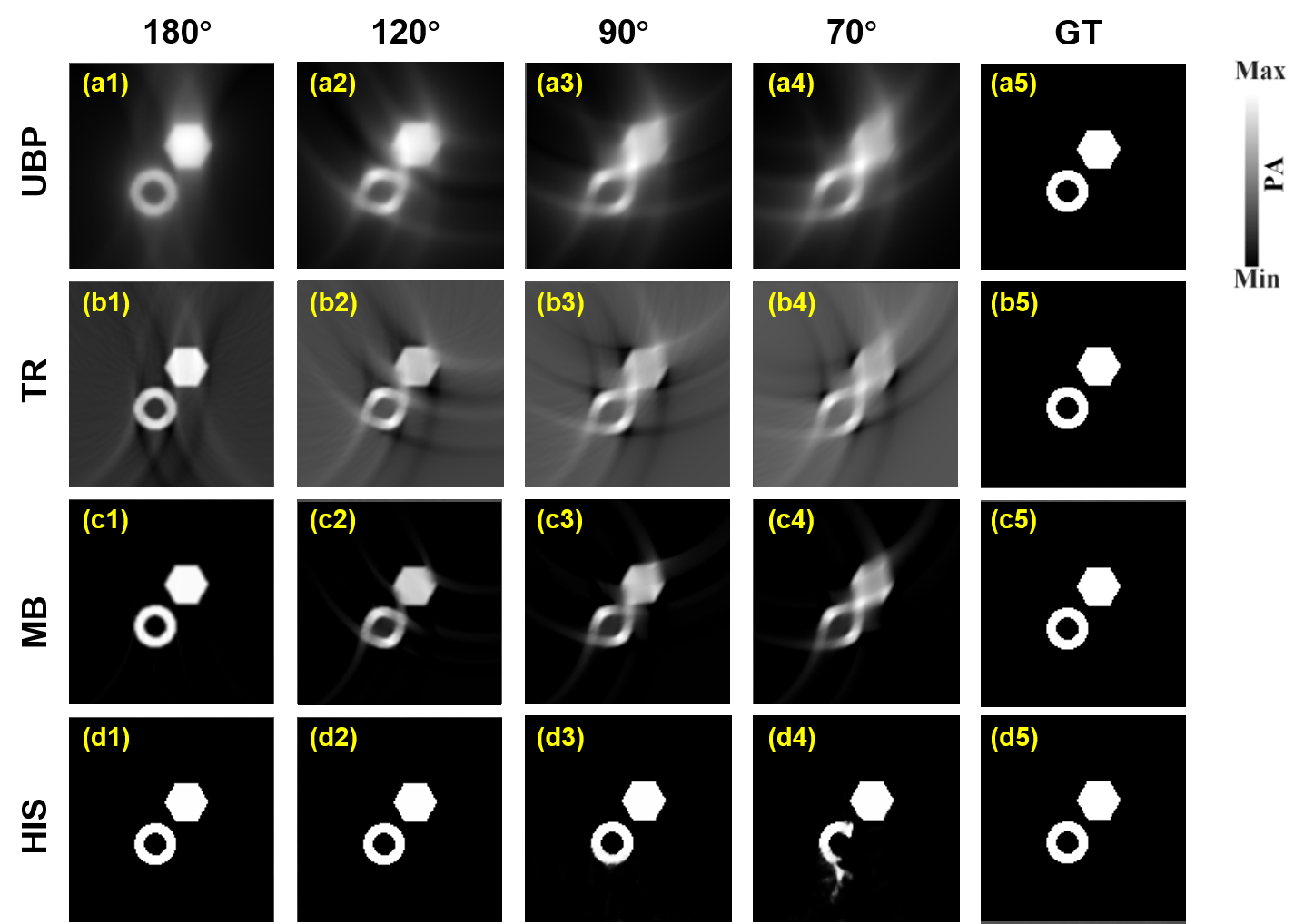}
    \caption{The reconstruction results of simple geometric phantom. (a1)-(a4) represent the results of the UBP method in limited-view cases of 180°, 120°, 90° and 70°, respectively. (b1)-(b4) are the results applying the TR method in limited-view cases of 180°, 120°, 90° and 70°, respectively. (c1)-(c4) show the results applying the MB method in limited-view cases of 180°, 120°, 90° and 70°, respectively. (d1)-(d4) represent the results of the HIS method in limited-view cases of 180°, 120°, 90° and 70°, respectively. (a5)-(d5) are the same ground truth. TR, time reversal. MB, model-based. GT, ground truth.}
    \label{fig:galaxy}
\end{figure}

\par Figure 5 presents the results for simulations involving a simplified vascular phantom. In this context, UBP, TR, and MB still manage to reconstruct the basic geometric shape of the simple vasculature fairly well at a 180° view, albeit with some noticeable artifacts present. As the view angle narrows to 120°, the reconstructions from UBP and TR become notably blurrier, with diminished detail. While MB retains more detail compared to the two, portions of the image begin to be lost. Further reduction to 90° and below sees all three conventional methods struggling to adequately reconstruct even the general layout of the simple vascular structure.
\par Conversely, the HIS method excels in preserving image detail, even under the extremely restrictive condition of a 70° view. It successfully reconstructs the majority of the intricate structures within the simple vasculature, highlighting its exceptional resilience to severe data limitations. This further solidifies the advantage of the HIS model in accurately recovering fine features under a wide range of limited-view, particularly those that pose significant challenges to conventional reconstruction techniques.

\begin{figure}[htp]
    \centering
    \includegraphics[width=9cm]{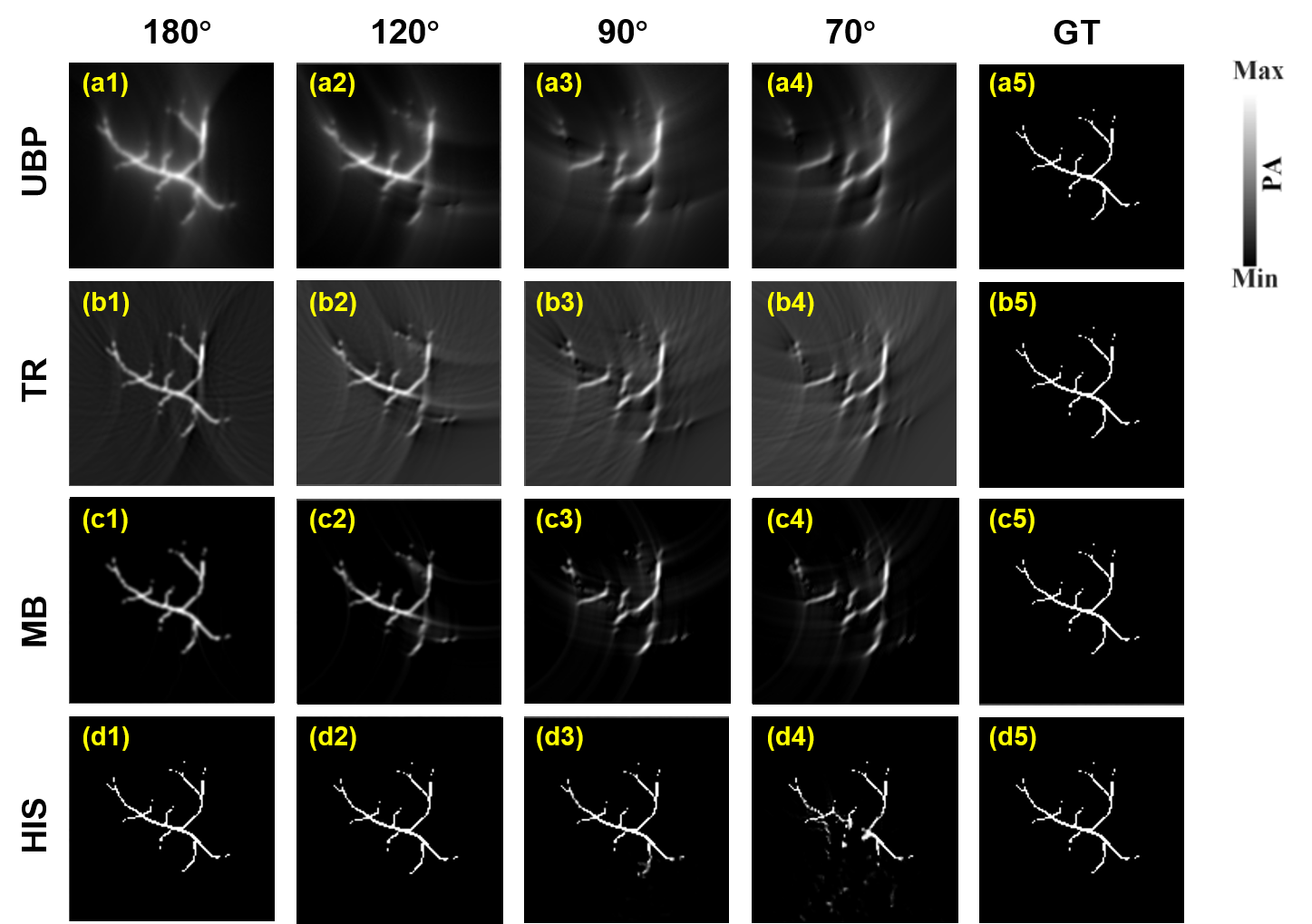}
    \caption{The reconstruction results of simplified vascular phantom. (a1)-(a4) represent the results of the UBP method in limited-view cases of 180°, 120°, 90° and 70°, respectively. (b1)-(b4) are the results applying the TR method in limited-view cases of 180°, 120°, 90° and 70°, respectively. (c1)-(c4) show the results applying the MB method in limited-view cases of 180°, 120°, 90° and 70°, respectively. (d1)-(d4) represent the results of the HIS method in limited-view cases of 180°, 120°, 90° and 70°, respectively. (a5)-(d5) are the same ground truth. TR, time reversal. MB, model-based. GT, ground truth.}
    \label{fig:galaxy}
\end{figure}

\par Figure 6 illustrates the outcomes for simulations using a complex vascular phantom. As the vasculature becomes more intricate, the reconstruction performance of UBP, TR, and MB significantly deteriorates, even when the view angles are relatively broad at 180° and 120°. Conversely, the HIS model continues to exhibit robust reconstruction capabilities, maintaining a high level of performance even with the increased complexity. This highlights the model's enhanced adaptability and resilience in handling sophisticated structures, further emphasizing its superiority over traditional methods when confronted with challenging, high-detail imaging tasks.
\par Table 1 summarizes the quantitative evaluation of the reconstructed results via Peak Signal-to-Noise Ratio (PSNR) and Structural Similarity Index (SSIM) metrics for the simulated experiments. Under the stringent condition of a 70° limited view, our proposed method achieved an average PSNR of 31.33 dB and an SSIM of 0.94, marking a substantial improvement over the Delay-and-Sum (DAS) method by 20.43 dB  respectively.


\begin{figure}[htp]
    \centering
    \includegraphics[width=9cm]{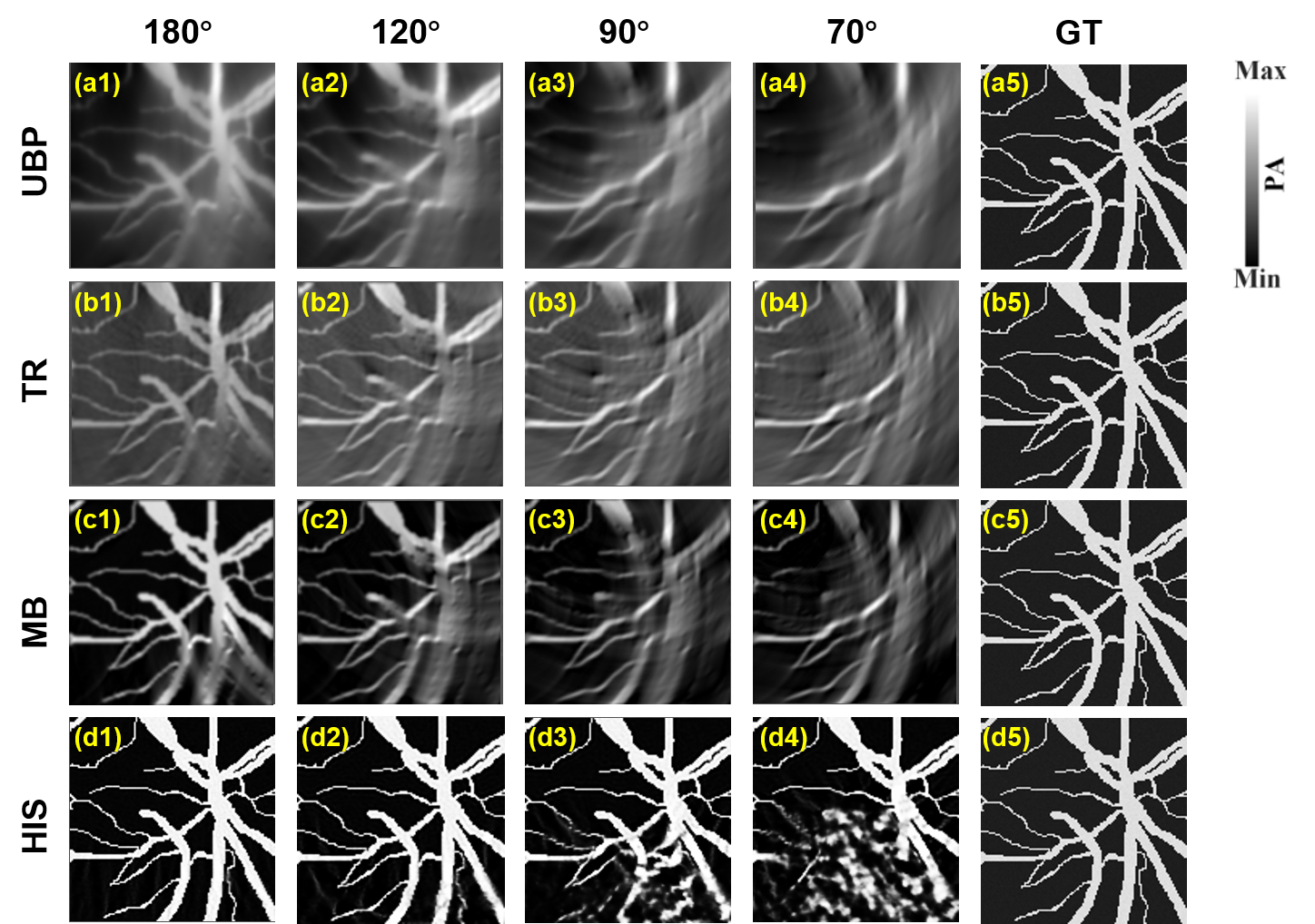}
    \caption{The reconstruction results of complex vascular phantom. (a1)-(a4) represent the results of the UBP method in limited-view cases of 180°, 120°, 90° and 70°, respectively. (b1)-(b4) are the results applying the TR method in limited-view cases of 180°, 120°, 90° and 70°, respectively. (c1)-(c4) show the results applying the MB method in limited-view cases of 180°, 120°, 90° and 70°, respectively. (d1)-(d4) represent the results of the HIS method in limited-view cases of 180°, 120°, 90° and 70°, respectively. (a5)-(d5) are the same ground truth. TR, time reversal. MB, model-based. GT, ground truth.}
    \label{fig:galaxy}
\end{figure}

\begin{table*}
\caption{\textbf{QUANTITATIVE RESULTS (PSNR/SSIM) OF ALL THE COMPARED METHODS ON SIMULATION DATA, PHANTOM EXPERIMENT DATA AND IN VIVO EXPERIMENTAL DATA FOR LIMITED VIEW OF 180°,120°,90°AND 70°}}
\centering
\begin{tabular}{cccccc}
\toprule
\textbf{Sample}&\textbf{Methods}&\textbf{180°}&\textbf{120°}&\textbf{90°}&\textbf{70°} \\
\midrule
\multirow{4}{*}{Simple geometric phantom}&UBP&14.64/0.0724&13.97/0.0568&13.90/0.0684&13.37/0.0718 \\
&TR&14.40/0.0451&10.90/0.0433&9.95/0.0240&9.13/0.0218 \\
&MB&25.78/0.9661&20.69/0.8306&18.90/0.7461&17.88/0.7656 \\
&HIS&35.83/0.9875&35.38/0.9831&35.24/0.9815&34.62/0.9793 \\
\midrule
\multirow{4}{*}{Simplified vascular phantom}&UBP&15.29/0.0646&15.78/0.0845&14.83/0.0217&15.10/0.0240 \\
&TR&15.37/0.0210&13.58/0.0324&12.83/0.0057&12.23/0.0041 \\
&MB&21.92/0.9243&20.56/0.8316&19.16/0.6404&19.03/0.6043 \\
&HIS&34.46/0.9955&34.15/0.9923&29.63/0.9831&23.77/0.8545 \\
\midrule
\multirow{4}{*}{Complex vascular phantom}&UBP&9.03/0.1179&8.69/0.1002&8.53/0.0988&8.47/0.0934 \\
&TR&9.49/0.1557&9.74/0.2196&8.15/0.0667&8.16/0.0640 \\
&MB&13.30/0.6218&10.91/0.4485&9.80/0.3391&9.28/0.2846 \\
&HIS&29.87/0.9696&26.47/0.9439&13.88/0.7870&11.15/0.6394 \\
\midrule
\multirow{4}{*}{Circular phantom data }&UBP&13.07/0.4175&15.62/0.7348&14.46/0.6795&10.69/0.6218 \\
&TR&15.76/0.7511&12.38/0.7354&10.93/0.6975&9.59/0.6187 \\
&MB&29.50/0.8928&20.41/0.8153&13.78/0.6579&12.38/0.6022 \\
&HIS&29.87/0.9696&26.47/0.9439&13.88/0.7870&12.50/0.6394 \\
\midrule
\multirow{4}{*}{In vivo experimental data}&UBP&10.90/0.3750&11.86/0.3656&11.32/0.3653&12.16/0.3877 \\
&TR&15.40/0.6730&9.48/0.5091&10.13/0.4087&11.00/0.4870 \\
&MB&20.22/0.7560&16.57/0.6815&14.65/0.6010&14.18/0.4861 \\
&HIS&33.21/0.9087&30.43/0.8550&24.31/0.6929&23.87/0.6360 \\
\bottomrule
\end{tabular}
\end{table*}

\subsection{Results on phantom experiment data}
To verify the efficacy of the proposed method on experimental data, we compared the reconstruction capabilities of HIS against UBP, TR, and MB methodologies using a circular phantom data, as illustrated in Figure 7. This figure exhibits the reconstruction outcomes derived from each of the four different methods under varying angular views of 180°, 120°, 90°, and 70°.

\par The proposed method can effectively reduce the artifacts in the image, even under the condition of a limited-view (e.g., 70°), a higher-quality reconstruction can be achieved than the other three methods. As the viewing angle increases, the reconstruction performance is further enhanced. Figure 7 illustrates these results, demonstrating that, with the exception of the extremely limited perspective at 70°, our proposed method successfully reconstructs the structure of the images and significantly suppresses the occurrence of artifacts. This highlights the high quality of the reconstructed cylindrical phantoms achieved by our proposed approach. Under the 180° scenario, the MB method effectively removes the majority of artifacts. Nevertheless, when confronted with highly limited angular view measurements, specifically at 70° and 90°, the reconstruction outcomes still exhibit some residual artifacts and a degradation in image structural clarity. As the angle decreases, there is an observed increase in image artifacts, leading to a decline in overall image quality.
\par The images reconstructed by the HIS method demonstrate more accurate detail preservation across varying degrees of limited-view. At 90°, the HIS method achieves a PSNR of 13.88 dB and a SSIM of 0.7870, marking an improvement over the MB method with an increase of 0.1 dB in PSNR and 0.1291 in SSIM. When the projection angle is reduced to 70°, the HIS method registers a PSNR of 12.50 dB and an SSIM of 0.6394, which are respectively 0.12 dB and 0.0372 higher than those achieved by the MB method. These experimental findings indicate that the HIS method outperforms the MB method, as well as the UBP and TR methods, in reconstructing cylindrical phantoms, particularly under conditions of exceedingly sparse projections.
\begin{figure}[htp]
    \centering
    \includegraphics[width=9cm]{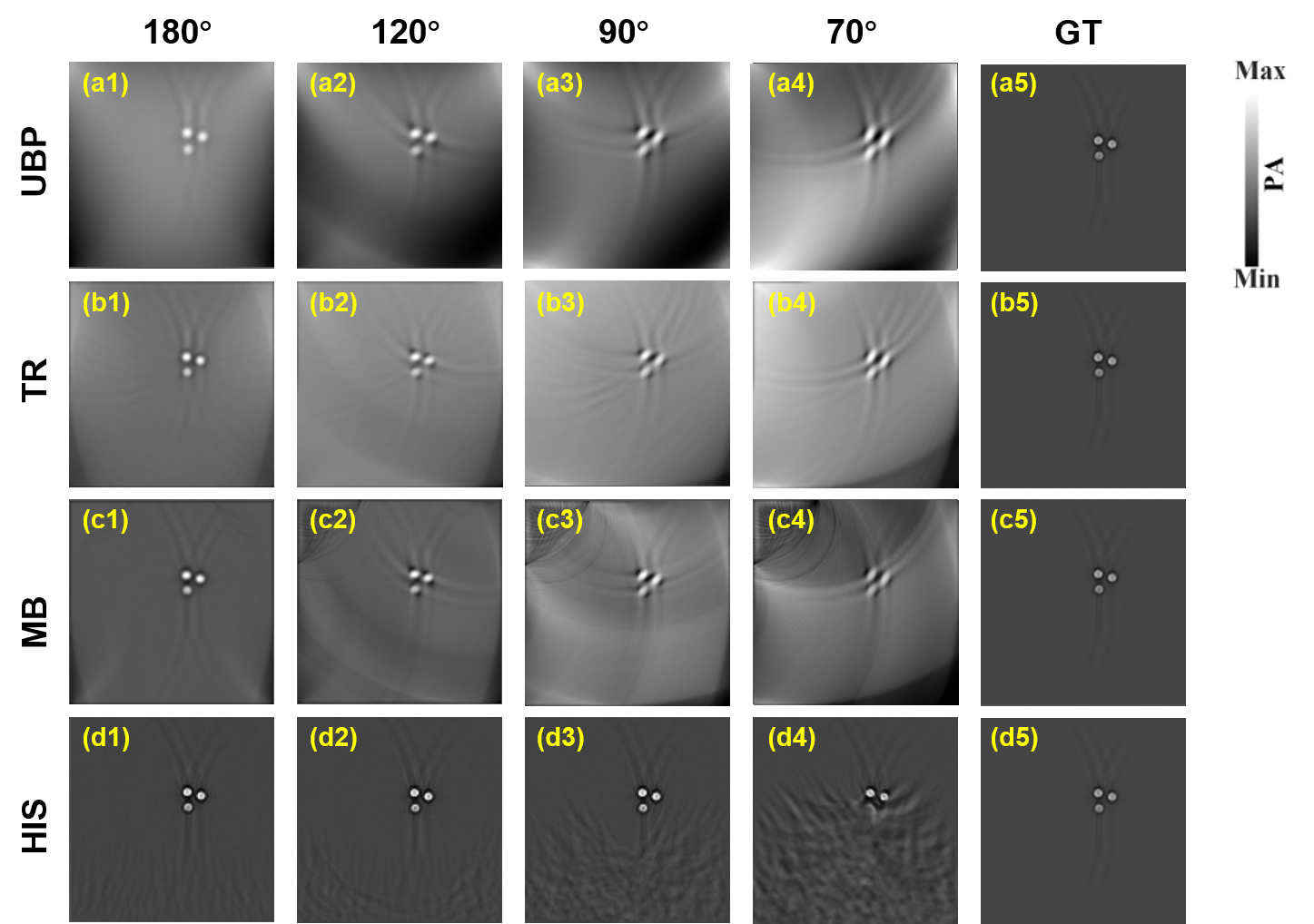}
    \caption{The reconstruction results of cicular phantom data. (a1)-(a4) represent the results of the UBP method in limited-view cases of 180°, 120°, 90° and 70°, respectively. (b1)-(b4) are the results applying the TR method in limited-view cases of 180°, 120°, 90° and 70°, respectively. (c1)-(c4) show the results applying the MB method in limited-view cases of 180°, 120°, 90° and 70°, respectively. (d1)-(d4) represent the results of the HIS method in limited-view cases of 180°, 120°, 90° and 70°, respectively. (a5)-(d5) are the same ground truth. TR, time reversal. MB, model-based. GT, ground truth.}
    \label{fig:galaxy}
\end{figure}

\subsection{Results on in vivo experiment data}
To further verify the applicability potential of the proposed method in limited-view PAT, subsequent experiments were conducted using in vivo data from a mouse abdomen. Figure 8 presents the reconstruction results obtained using the UBP method, the TR method, the MB method, and the HIS method for comparison.
\par Table 1 summarizes the quantitative results of the in vivo experiments in terms of PSNR and SSIM, further corroborating the superiority of the proposed method. Under a 180° limited field of view setting, HIS yields notably higher PSNR and SSIM values, exhibiting an improvement of 12.99 dB in PSNR and 0.1527 in SSIM over the MB method. In the case of the extreme limited-angle scenario at 70°, the HIS method achieves a PSNR of 23.87 dB and an SSIM of 0.6360, marking respective enhancements of 9.69 dB and 0.1499 compared to the MB method. These metrics quantitatively reinforce the enhanced reconstruction capability of HIS, particularly under severely constrained angular views.The experimental results indicate the HIS possesses exceptional performance on boosting image contrast and eradicating artifacts in extremely restrict limited-view cases.

\begin{figure}[htp]
    \centering
    \includegraphics[width=9cm]{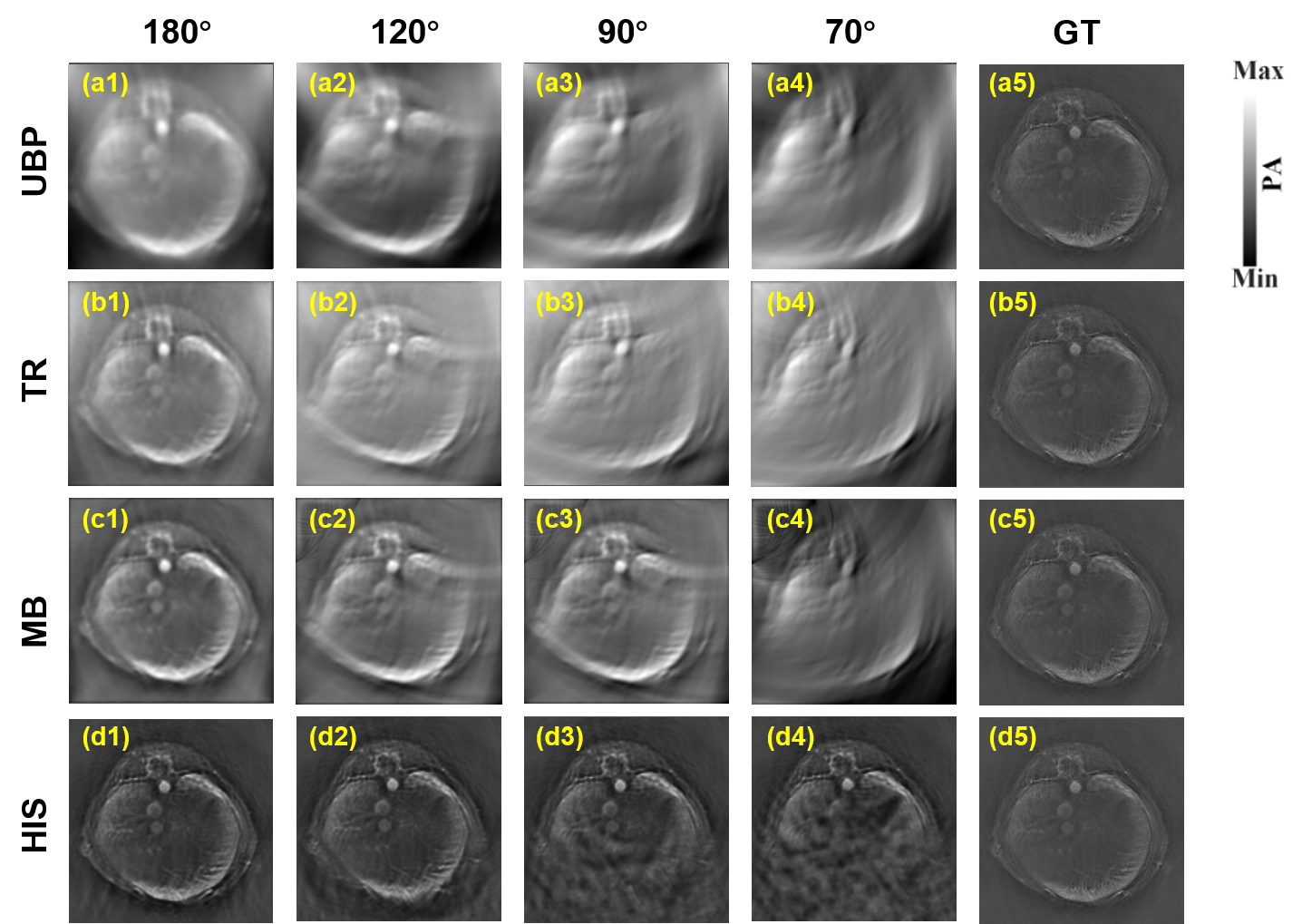}
    \caption{The reconstruction results of in vivo mouse data. (a1)-(a4) represent the results of the UBP method in limited-view cases of 180°, 120°, 90° and 70°, respectively. (b1)-(b4) are the results applying the TR method in limited-view cases of 180°, 120°, 90° and 70°, respectively. (c1)-(c4) show the results applying the MB method in limited-view cases of 180°, 120°, 90° and 70°, respectively. (d1)-(d4) represent the results of the HIS method in limited-view cases of 180°, 120°, 90° and 70°, respectively. (a5)-(d5) are the same ground truth. TR, time reversal. MB, model-based. GT, ground truth.}
    \label{fig:galaxy}
\end{figure}

\section{Conclusion and discussion}
In this work, we propose a novel deep learning framework, named HIS, based on implicit representation to enhance the quality of photoacoustic image reconstruction under limited viewing angles. In HIS, we integrate spatial encoding with fully connected neural networks, training a robust model to precisely predict photoacoustic image intensity (initial pressure) from input spatial coordinates. Reconstruction results on multiple sets of simulated images with varying limited viewing angles demonstrate HIS's ability to accurately approximate implicit spatial representations. Furthermore, compared to supervised CNN-based methods, HIS doesn’t requires training data, enhancing its generalization capabilities.


%

\ifCLASSOPTIONcaptionsoff
  \newpage
\fi



%
\bibliographystyle{unsrt}
\bibliography{refs}

%
\end{document}